\documentclass[10pt,journal,compsoc]{IEEEtran}

\ifCLASSOPTIONcompsoc
  \usepackage[nocompress]{cite}
\else
  \usepackage{cite}
\fi

\ifCLASSINFOpdf
  \usepackage[pdftex]{graphicx}
\else
\fi

\newcommand{\etal}{\textit{et al.}}
\usepackage{amsmath}
\usepackage{amssymb}
\usepackage{multirow}
\usepackage{subfigure}
\usepackage{color}

\usepackage{array}
\usepackage{setspace}
\usepackage{arydshln}
\usepackage{booktabs} 

\usepackage{soul,color}
\definecolor{Red}{cmyk}{0,1,1,0}
\definecolor{Green}{cmyk}{1,0,1,0}
\definecolor{Cyan}{cmyk}{1,0,0,0}
\definecolor{Purple}{cmyk}{0.45,0.86,0,0}
\definecolor{Rosolic}{cmyk}{0.00,1.00,0.50,0}
\definecolor{Blue}{cmyk}{1.00,1.00,0.00,0}
\definecolor{orange}{cmyk}{0,0.52,0.80,0}
\definecolor{Orange_light}{cmyk}{0,0.2,0.86,0}
\definecolor{Black}{cmyk}{1,0,0,1}

\newcommand{\gl}[1]{{\color{black}#1}}

\newcommand{\yt}[1]{{\color{black}#1}}

\newcommand{\ytt}[1]{{\color{black}#1}}
\newcommand{\ytFinal}[1]{{\color{black}#1}}

\begin{document}

\title{{\huge Robust Pose Transfer with Dynamic Details using \quad Neural Video Rendering}}

\author{Yang-Tian Sun, Hao-Zhi Huang, Xuan Wang, Yu-Kun Lai, Wei Liu, and Lin~Gao\IEEEauthorrefmark{1}
\thanks{\IEEEauthorrefmark{1} Corresponding Author is Lin Gao (gaolin@ict.ac.cn).}
\IEEEcompsocitemizethanks{
\IEEEcompsocthanksitem Y.-T. Sun and L. Gao are with the Beijing Key Laboratory of Mobile Computing and Pervasive Device, Institute of Computing Technology, Chinese Academy of Sciences, Beijing, China, and also with the University of Chinese Academy of Sciences, Beijing, China.\protect\\
E-mail:sunyangtian@ict.ac.cn, gaolin@ict.ac.cn
\IEEEcompsocthanksitem H.-Z. Huang is with the Xverse.
\protect\\
E-mail: huanghz08@gmail.com

\IEEEcompsocthanksitem X. Wang and W. Liu are with the Tencent. 
\protect\\
E-mail: xwang.cv@gmail.com, wl2223@columbia.edu

\IEEEcompsocthanksitem Y.-K. Lai is with the  School of Computer Science and Informatics, Cardiff University, Wales, UK.
\protect\\
E-mail:LaiY4@cardiff.ac.uk

}
}

\markboth{IEEE Transactions on Pattern Analysis and Machine Intelligence,~Vol.~xx, No.~xx, June~2021}%
{Lin \MakeLowercase{\textit{et al.}}: Robust Pose Transfer with Dynamic Details using Neural Video Rendering}

\IEEEtitleabstractindextext{%
\begin{abstract}
Pose transfer of human videos aims to generate a high-fidelity video of a target person imitating actions of a source person. A few studies have made great progress either through image translation with deep latent features or neural rendering with explicit 3D features. 
However, 
both of them rely on large amounts of training data to generate realistic results, and the performance degrades on more accessible Internet videos due to insufficient training frames.
In this paper, we demonstrate that the dynamic details can be preserved even when trained from short monocular videos.
Overall, we propose a neural video rendering framework coupled with an image-translation-based dynamic details generation network (D$^2$G-Net), 
which fully utilizes both 
the stability of explicit 3D features
and the capacity of learning components.
To be specific, 
a novel hybrid texture representation is presented to encode both the static and pose-varying appearance characteristics, 
which is then mapped to the image space and rendered as a detail-rich frame in the neural rendering stage.
Through extensive comparisons, we demonstrate that our neural human video renderer is capable of achieving both clearer dynamic details and more robust performance \yt{even on accessible short videos with only 2k$\sim$4k frames.}
\end{abstract}

}

\maketitle

\begin{figure*}[t]
\begin{center}  
\includegraphics[width=0.95\textwidth]{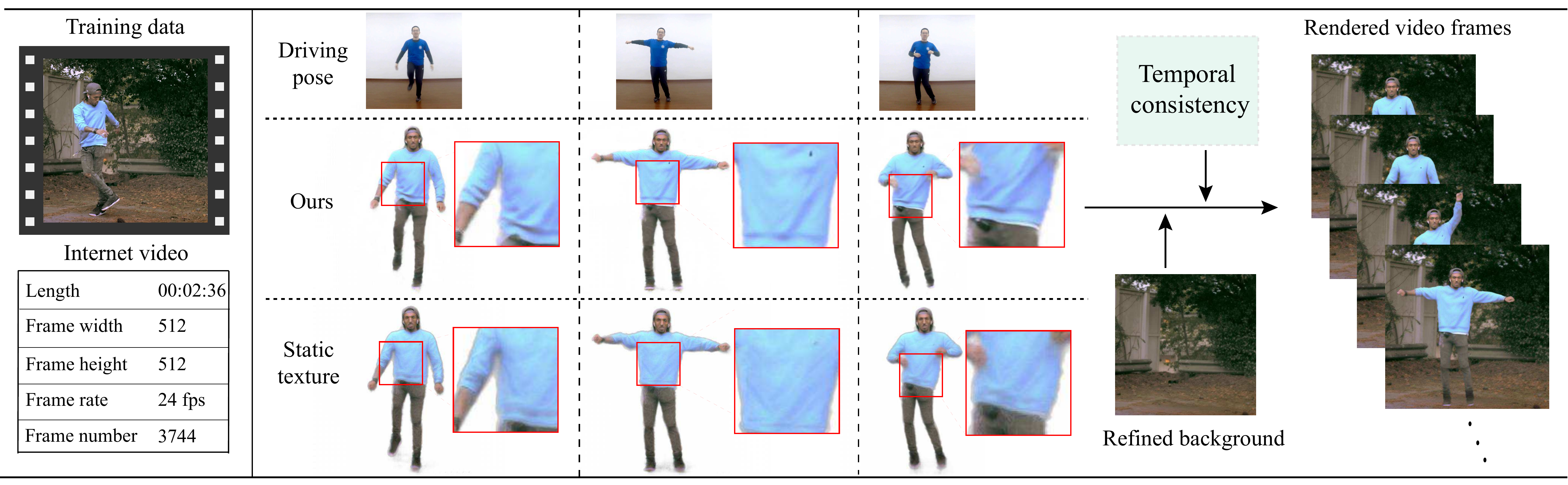}
\end{center}
\vspace{-3mm}
\caption{With only an accessible Internet video as training data, our neural rendering framework is able to synthesize temporally coherent person images from given pose sequences, while keeping rich details compared with traditional rendering results using static textures.}
\vspace{-6mm}
\label{fig:teaser}
\end{figure*}

\IEEEdisplaynontitleabstractindextext

%
\IEEEpeerreviewmaketitle

\IEEEraisesectionheading{\section{Introduction}\label{sec:introduction}}

\IEEEPARstart{R}{ecently}, great progress has been achieved by applying neural networks to video synthesis, especially the human motion transfer task, which aims to transfer the action of a source person depicted in a video to a target one. The most essential challenges for this problem include but are not limited to synthesizing a detail-rich target video and imitating a variety of human motions, 
which can differ significantly from the training set.

Most existing approaches fall into one of the two categories: image-to-image translation methods~\cite{everybody, vid2vid, Dense_Intrinsic_Appearance_Flow, Deep_Motion_Transfer,PG, First_Order_Motion_Model} which map pose labels to person images,
and neural rendering based techniques that learn explicit 3D representations coupled with the traditional graphics rendering pipeline. Despite the powerful representational capacity of deep features that can help generate detail-rich results, image-to-image translation methods rely on black-box 2D generative networks and often introduce obvious artifacts for poses with a large deviation from the training samples. Hence, a great number of training frames are needed to cover as many poses as possible, e.g., more than 10k frames are used for each subject in \cite{everybody}. For neural rendering based techniques, the explicit 3D representations are introduced and hence the generative models become more stable. However, they either learn a static texture representation during the training stage, e.g., ~\cite{TNA, densepose_transfer}, which is temporally invariant and often leads to blurry results, or rely on finely reconstructed 3D models~\cite{dynamic_texture}, which are difficult to obtain without large amounts of multi-view data.
In general, all of these methods degrade when training data is limited to a common monocular video, which is more accessible in our daily life, e.g., short videos from the Internet.

To alleviate the aforementioned problems, we propose a novel generative framework that seamlessly couples image translation components and neural rendering, to gain the benefits of both and works well even with only a short monocular video. 
\yt{We ease the training difficulty and reduce the data dependency by presenting a novel learnable texture representation together with a corresponding pose-aware \emph{dynamic details generation network} (D$^2$G-Net) embedded in the neural renderer.}
Specifically, rather than using a traditional static RGB-channel image, we represent the texture with a hybrid feature map, 
\yt{which encodes the static RGB colors explicitly and the dynamic details of human appearance implicitly.}
Therefore, the hybrid texture serves as a higher level description of human appearance compared with classic texture images.
Meanwhile, we also predict the texture coordinates of each pixel in each frame directly for more flexible
UV mapping, \yt{which sidesteps the difficulty of fine 3D reconstruction.}
\yt{By this means, the learned hybrid texture can be adaptively mapped to the 2D screen space in a differentiable way without the traditional rasterization rendering pipeline,
\yt{which is then fed into D$^2$G-Net for dynamic detail generation. Since the human appearance details are pose-varying, the generation process of D$^2$G-Net is also conditioned on the current pose.}
Hence, the mapped feature can be rendered into a foreground human figure with clear details \emph{dynamically}.}
\yt{Moreover, to obtain the complete background image, we propose a refinement strategy by integrating the background information from all the video frames with different foreground-background occlusions due to different poses.} Finally, we composite the predicted foreground and background together to obtain the final rendered result. 
\ytt{
Note that simply splicing the individually generated frames to the video 
introduces flickers and jitters. To deal with
such artifacts, we incorporate the temporal loss~\cite{temporal_loss} 
to synthesize temporally consistent videos.}
Here, we summarize the \ytt{technical} contributions as follows: 
\begin{itemize}
\item \yt{A novel end-to-end neural rendering framework for human video generation with dynamic details using 
accessible monocular video as training data.}
\item \ytt{A learnable hybrid texture representation 
to encode static appearance explicitly and high-frequency pose-varying details implicitly.}
\item \yt{A pose-aware dynamic detail generation network (D$^2$G-Net) which renders the learned texture feature into high-fidelity human images with pose-varying details.}
\end{itemize}

\vspace{-5mm}
\section{Related Work}
Video-based human motion transfer has been extensively studied over the recent decades due to its ability for fast video content production. The early approaches accomplished this task by manipulating existing video footage, e.g., Video Rewrite~\cite{video_rewrite}. 
\ytt{Benefiting from the rapid development of deep learning, both Generative Adversarial Networks (GANs)~\cite{cGAN,cycleGAN2017,discoGAN} and neural rendering techniques~\cite{thiesDeferred2019, dynamic_texture, TNA, lombardiDeep2018, liuNeural2019} are applied to the motion transfer task.} \ytt{Here we summarize major advances in the two branches of research, followed by an overview of the methods dedicated to video
synthesis.}

{\bf Image-translation based Approaches.} \ytt{
Image-to-image translation was proposed in Pix2pix~\cite{pix2pix}, where a conditional GAN with a U-net architecture~\cite{unet} was used to transfer an image from one domain to another. Pix2pixHD~\cite{pix2pixHD} introduced a multi-scale generator to the Pix2pix pipeline for high-resolution image synthesis. Such architectures and training methods were applied to the video-based pose transfer task and achieve visually pleasing results~\cite{everybody, vid2vid}. However, the fully-supervised learning can easily cause overfitting, leading to poor results when the desired poses are quite different from the training set. There are also other approaches with elaborately-designed pipelines based on the image translation architecture for more general pose-guided person image synthesis, e.g.,~\cite{PG, Dense_Intrinsic_Appearance_Flow, Deep_Motion_Transfer,First_Order_Motion_Model, LWGAN, posewarp, densepose_transfer, Siarohin2018DeformableGF,Esser2018AVU, Ma2018DisentangledPI}. 
Esser \etal~\cite{Esser2018AVU} decoupled the appearance from the shape-guided image synthesis network, enriching the control over the image-translation network. \cite{Dense_Intrinsic_Appearance_Flow, Deep_Motion_Transfer,First_Order_Motion_Model, LWGAN, Siarohin2018DeformableGF} assisted the pose-guided generation with delicate warping strategies on images or feature maps. 
However, these methods focus on the synthesis of general human images with low resolution, while ours aims to generate high-fidelity person-specific video.
}

{\bf Human Image Synthesis based on Neural Rendering.} Some approaches performed human body synthesis by employing an explicit 3D representation and neural rendering. 
Textured Neural Avatars~\cite{TNA} learned a full-body neural avatar by estimating an explicit texture map and mapping the input pose to a UV-coordinate image. As \emph{static} texture maps are used, most high-frequency details are consequently lost in the synthesized results. Liu \etal~\cite{dynamic_texture} also proposed to predict dynamic texture depending on poses. However, they employed accurate 3D reconstructions captured by dedicated devices, which are not practical in most application scenarios. 
\ytt{In this paper, a novel hybrid texture representation 
with a dynamic detail generation network is proposed for the realistic and detail-rich person image generation even without high-accuracy 3D reconstructions.}

{\bf Video Synthesis.} 
\ytt{
Early efforts on video synthesis attempted to improve continuity in the time domain via a recurrent neural network~\cite{videoGAN, mocoGAN}. Wang \etal~\cite{vid2vid} appended an optical flow prediction module and a video discriminator to the off-the-shelf image translation network~\cite{pix2pixHD}.  Here we introduce the temporal loss from~\cite{temporal_loss} to enforce the network to synthesize temporally consistent results inherently without additional network modules.
}



\begin{figure*}[t]
\begin{center}
\includegraphics[width=0.95\textwidth]{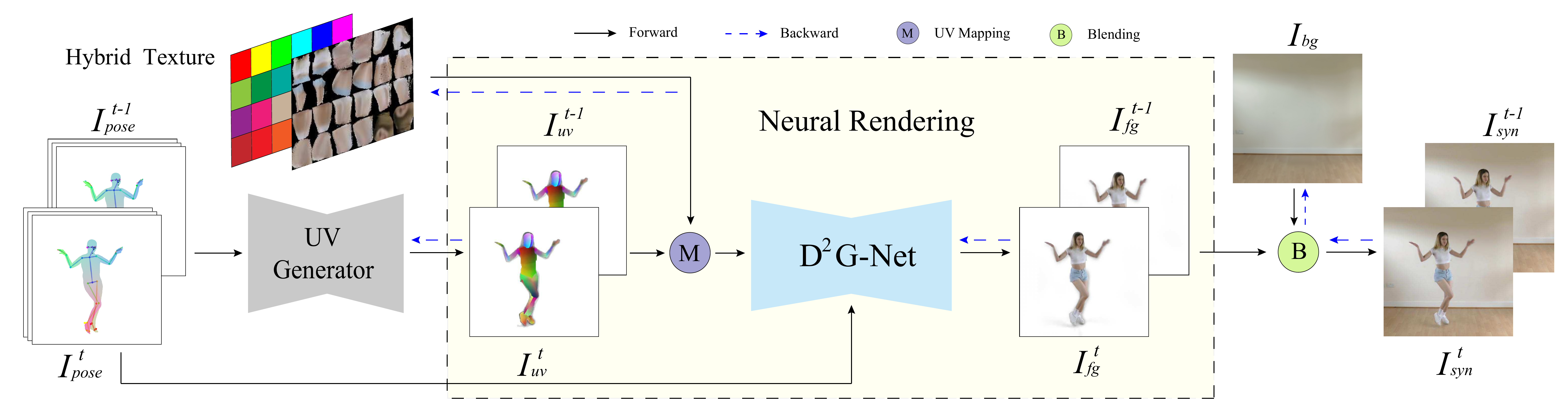}
\end{center}
\vspace{-6mm}
\caption{The pipeline of our training process. Pose labels $I_{\text{pose}}^{t-1}$ and $I_{\text{pose}}^t$ are extracted from \ytt{the temporal context of} adjacent video frames. 
\yt{The UV generator takes pose labels as input and predicts the UV coordinates of each pixel, i.e., $I_{\text{uv}}$. Afterwards, the hybrid texture is mapped to the screen space according to the predicted UV coordinates and then translated to human foreground images with dynamic details.  }
Meanwhile, background image is refined during the training process and final synthesized images are obtained through a combination of foreground and background images. 
\yt{Note that adjacent frames are trained as a whole for the implicit learning of temporal coherence in the neural video renderer.}}
\label{fig:pipeline}
\vspace{-3mm}
\end{figure*}

\vspace{-5mm}
\section{Method}
\subsection{Overview}
Our aim is to generate a new video of the target person imitating the specific movements \yt{using only a short monocular video as training data}, while keeping high fidelity and temporal consistency. To achieve this, we propose a novel neural video rendering framework containing both \yt{a hybrid texture representation and an embedded dynamic details generation network}, as illustrated in Fig.~\ref{fig:pipeline}.
\yt{Specifically, \ytt{given extracted pose labels,} we first predict the texture coordinates of each pixel through a UV generator. Meanwhile, we initialize the texture with a learnable feature map. Afterwards, based on the predicted UV coordinates the texture feature is mapped to the screen space, which is then fed into the pose-conditioned dynamic details generation network for detail-rich human foreground image rendering.}
\yt{Moreover, to complete the generated frame we propose to combine the background refinement into the end-to-end training framework, by which means the information of frames with different foreground-background occlusions can be integrated effectively.}

\vspace{-3mm}
\subsection{Foreground Image Generation}
In the classical graphics rendering pipeline, the texture maps are always static RGB images and not conducive to the characterization of high-frequency detailed signals, while the geometric deformations are responsible for dynamic details.
DTL~\cite{dynamic_texture} first proposes to learn a dynamic texture to characterize the high-frequency signals of human appearance. However, its dynamic texture learning is under the supervision of unwrapped video frames, which relies on the fine-grained character model reconstructed from multi-view training data.
\gl{Here, we propose a novel neural rendering framework to enrich the dynamic fidelity details with a} hybrid texture representation and the corresponding dynamic detail generation network (denoted as ``D$^2$G-Net``). The former contains both explicit static color and implicit details features of human appearance, while the latter is responsible for varying signals visualization dynamically in the neural rendering process.
\yt{Here we start introducing our pipeline from the representation of pose labels, followed by the constitution of hybrid texture representation. Then we expound the neural rendering process, which includes the UV mapping as well as the dynamic details generation from texture features.} 

{\bf Pose Labels Representation.} \ytt{Our pose labels contain both 2D and 3D features.} On the one hand, the 2D pose label is a skeleton image based on keypoints extracted by OpenPose~\cite{openpose}. On the other hand, the 3D label is the projection image of a 3D human mesh, which is obtained using a video-based reconstruction approach~\cite{total_capture}. Each pixel of the 3D labels contains 3-channel Laplacian features~\cite{human_motion_tranfer_3d}, which are intrinsic \yt{characterization of 3D geometry} and capture 3D body shapes meanwhile. We adopt 2D keypoints because they are relatively more accurate and easy for tracking, while 3D information can cope with pose ambiguity and self-occlusion. \ytt{The 2D and 3D labels are concatenated into a 6-channel image to represent the current pose. Considering the effect of pose trajectory on dynamic details generation, we expand the current pose label by introducing its temporal context, i.e., we form the pose label as the concatenation of the current frame pose and last two frames. 
\ytFinal{Such a trajectory-based pose label is conducive to more realistic dynamic characteristics generation due to the consideration of velocity and acceleration, as we report in the supplementary material.}
}

\begin{figure}[t]
\begin{center}
\includegraphics[width=\linewidth]{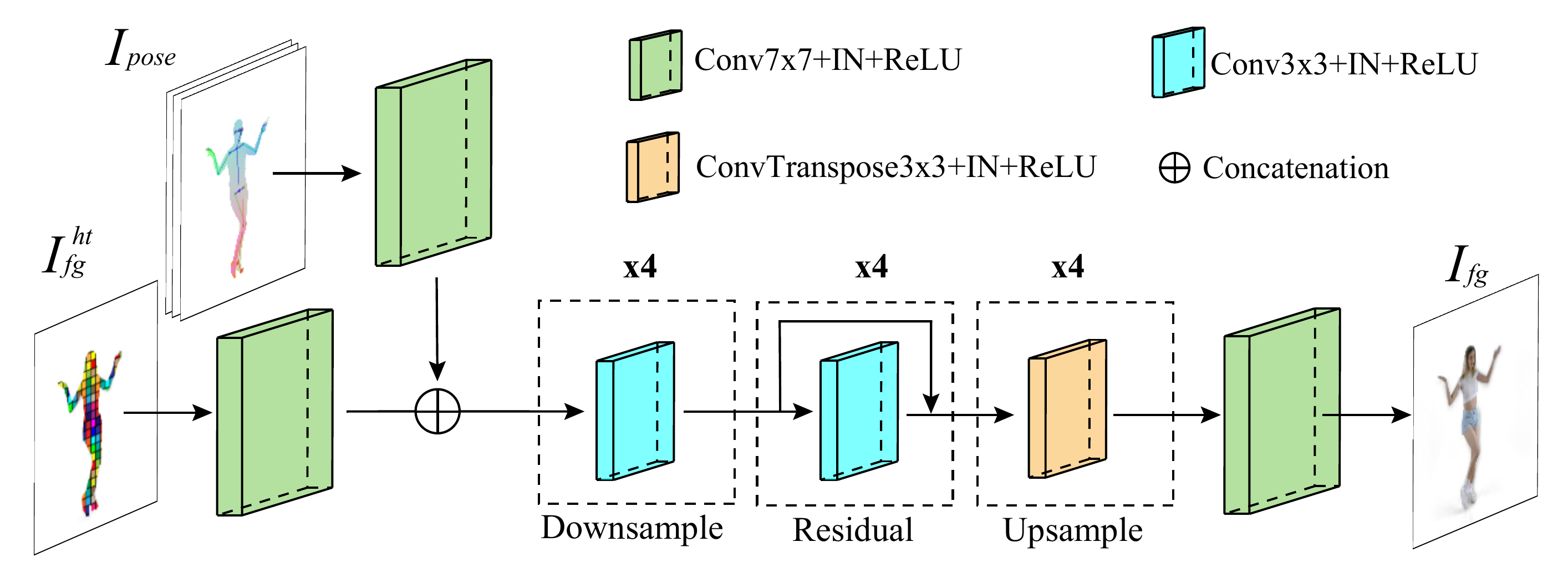}
\end{center}
\vspace{-3mm}
\caption{\yt{The illustration of D$^2$G-Net. $I_{\text{fg}}^\text{ht}$ is obtained by mapping the hybrid texture to the screen space. The translation network aims to translate the high-dimensional feature to RGB color with dynamic details under the current pose. Note that the pose label is conditioned on the feature level to guide the translation process.}}
\vspace{-6mm}
\label{fig:tex_gen}
\end{figure}

{\bf Hybrid Texture Representation and Initialization.}
\yt{To alleviate the decrease of fidelity caused by static texture map, we introduce hybrid texture representation, which is a high-dimensional feature map consisting of both explicit RGB color and implicit detail characteristics. The hybrid texture enriches the generated details by 1) encoding the appearance details implicitly and compensating the explicit RGB map and 2) regularizing the UV generator during the joint learning with D$^2$G-Net.}
\yt{Technically, the texture formation follows the DensePose system~\cite{densepose}, which unwraps the human body into $N$($N=24$) patches and provides a mapping from each pixel to a certain patch of the human surface.}
\yt{Therefore, the hybrid texture is also composited of $N$ parts (denoted as $T_i$, where $i=1,\cdots,N$), corresponding to $N$ parts of human body.
The hybrid texture contains 18 channels,
\ytt{of which the first three are initialized as RGB colors by unwrapping each frame to the texture space according to DensePose, while the others are initialized as zero.}
\ytt{We choose the DensePose instead of the reconstructed 3D model for texture initialization due to its better alignment with frames and clearer results. We provide a more detailed explanation of our design choices in the supplementary material.} 
}

{\bf Neural Rendering.} 
\yt{In this stage, we map the hybrid texture to the screen space differentiably by predicting the UV coordinate through \ytt{a} UV generator. 
The mapped features are then fed into D$^2$G-Net for detail-rich human foreground generation.}

\textit{UV Generation.}
The traditional rendering pipeline relies on explicit 3D models for texture mapping. However, it is difficult to obtain fine 3D human models from only a monocular video. \yt{Moreover, the rasterization blocks the backward gradient since its indifferentiable characteristic, which makes the end-to-end training impossible.} In our pipeline, we resolve this problem by predicting texture coordinates (UV) of each pixel in each frame directly with a UV generator. 
To be specific, the UV generator takes current pose labels as input, and outputs the UV coordinates and part probabilities for each pixel in the video frame. The part probabilities ${P}_i (i=0,1,\cdots,N)$ have the size $H \times W$, representing the probability of each pixel belonging to $N$ parts (${P}_1,\cdots,{P}_N$) or background (${P}_0$), while the coordinates ${C}_i (i=1,\cdots,N)$ have the size of $H \times W \times 2$, indicating the UV coordinates of each pixel in the corresponding part. Here $H$ and $W$ are the height and width of video frames, respectively. Then the human foreground of \textit{hybrid} \textit{texture} $I_{\text{fg}}^{\text{ht}}$ can be obtained by
\begin{equation}
    I_{\text{fg}}^{\text{ht}} = \sum_{i=1}^N {P}_i \cdot \phi(T_i, {C}_i),
\end{equation}
where $\phi$ is a function that maps the hybrid texture $T_i$ to the screen space according to UV coordinates $C_i$.
\yt{We extract the static component, i.e., RGB channels from $I_{\text{fg}}^{\text{ht}}$ as $\widetilde{I}_{\text{fg}}$, which is a human foreground image without details and serves for subsequent regular loss calculation (Eq.~\ref{equa:supervised}).}
By this means we avoid
explicit 3D shape modeling \yt{and provide a way for end-to-end training, which enables us to find the best hybrid texture representation and corresponding D$^2$G-Net.}

\yt{
\textit{Dynamic details generation.} 
In this part, we aim to visualize the implicit details contained in texture features by translating the human foreground of hybrid texture, i.e., $I_{\text{fg}}^{\text{ht}}$, into detail-rich human images, denoted by $I_{\text{fg}}$.
Since the same texture feature should be interpreted to different appearance characteristics under different poses (e.g., wrinkles on the belly when bending down while flat clothes when standing erectly), we introduce the pose label as guidance during the dynamic details generation process. 
We adopt the pix2pixHD\cite{pix2pixHD} generator as the backbone of our D$^2$G-Net, which has achieved excellent results on the image-translation tasks. The pose label $I_{\text{pose}}$ is conditioned by concatenating with $I_{\text{fg}}^{\text{ht}}$ at the intermediate level, as illustrated in Fig.~\ref{fig:tex_gen}. 
The functionality of D$^2$G-Net can be formatted as 
\begin{equation}
    I_{\text{fg}} = \text{D}^2\text{G}(I_{\text{fg}}^{\text{ht}}, I_{\text{pose}}) .
\end{equation}
Note that although $I_{\text{fg}}^{\text{ht}}$ also contains pose information, the varying \ytt{features} increase the difficulty of \ytt{convergence} during training. We perform an ablation study of pose condition in Sec.~\ref{subsec:albation_image_translation}.
}

\vspace{-4mm}
\subsection{Background Refinement and Combination}
We will complete the generated detail-rich human foreground into a final video frame in this subsection. \yt{Since we focus on the human video generation with static background, we optimize the background image during the training process, which utilizes the information from all frames. In order to start from a reasonable initial state, we initialize the background image with a state-of-the-art inpainting network.}

We obtain the coarse initial background through frame-by-frame human body deduction and inpainting. Specifically, we segment the foreground (human) based on a U-net~\cite{unet} from the background for each frame in the training video, and then average the inpainted results obtained using the image inpainting approach~\cite{deepfillv2}.

\yt{During the training stage, the initial background image is updated according to the backpropagated gradient. By this means the information of all frames is aggregated and the inpainted artifacts are eliminated effectively, as illustrated in \ytt{the supplementary material.}}
Since ${P}_0$ denotes the probability of each pixel belonging to the background, we can combine the foreground and background by
\begin{align}
    I_{\text{syn}} &= I_{\text{fg}} \odot (1 - {P}_0) + I_{\text{bg}} \odot {P}_0 ,\\
    \widetilde{I}_{\text{syn}} &= \widetilde{I}_{\text{fg}} \odot (1 - {P}_0) + I_{\text{bg}} \odot {P}_0,
\end{align}
where $I_{\text{syn}}$ is the final synthesized frame, $\widetilde{I}_{\text{syn}}$ is the synthesized frame of the static component, i.e., RGB channels from the hybrid texture, and $\odot$ represents element-wise multiplication.

\vspace{-11mm}
\subsection{Temporal Consistency}
\vspace{-3mm}
The video produced by processing each frame individually does not look realistic and natural enough because of the inevitable flickering and jittering artifacts, especially when the dynamic details are presented. 
To address this problem, we introduce the temporal loss~\cite{temporal_loss} into the human video generation task, which is defined as the  $L_1$ loss between the generated frame $I_{\text{syn}}^t$ at time $t$ and the warped version of the generated frame
$I_{\text{syn}}^{t-1}$ at time $t-1$. 
\ytt{The detailed definition and effect of the temporal loss are presented in the supplementary material.}

\vspace{-11mm}
\subsection{Full Objective}
\vspace{-3mm}
We denote the UV generator and D$^2$G-Net as $G_{\text{uv}}$ and $G_{\text{D}^2\text{G}}$, respectively. 
\ytt{Note that $G_{\text{uv}}$ is first pretrained on frames of a large corpus of different characters, which can be used for the end-to-end training of other characters.}
\ytt{The pretraining helps improve the model robustness, and the finetuning process enables the generated UV map to better fit the specific character's body shape, as we demonstrate in the supplementary material.}

The pretraining objective is given by minimize the following loss
\begin{equation}
    \mathcal{L}_{\text{uv}}(P, \hat{P}, C, \hat{C}) = \mathcal{L}_{CE}(P, \hat{P}) + || C - \hat{C} ||_1 ,
\end{equation}
where $\mathcal{L}_{CE}$ is the cross-entropy loss. $P$ and $C$ are the predicted probabilities and UV coordinates of the UV generator, while $\hat{P}$ and $\hat{C}$ are the ground truths. Note that $\hat{P}$ and $\hat{C}$ can be obtained through existing 3D priors, e.g., rasterization of rigged 3D human models. In practice, we use the results of DensePose directly.

\yt{Our whole training network can be trained in an end-to-end manner. Let $I_{\text{pose}}$, $I_{bg}$ $I_{\text{real}}$ and $T$ be the pose label, background image, ground truth image, and hybrid texture, respectively. The overall objective is formulated as:}
\begin{align}
\nonumber  \min_{\mbox{\tiny$\begin{array}{c}
  G_{\text{D}^2\text{G}},G_{\text{uv}} \\
  I_{\text{bg}}, T \end{array}$}}
& ((\max\limits_{D} \sum_{i \in [t-1,t]} \mathcal{L}_{\text{GAN}}(I_{\text{pose}}^i, I_{\text{syn}}^i, I_{\text{real}}^i)) \\
\nonumber &+ \sum_{i \in [t-1,t]} \mathcal{L}_{\text{supervised}}(I_{\text{syn}}^i, I_{\text{real}}^i) \\
\nonumber &+ \sum_{i \in [t-1,t]} \mathcal{L}_{\text{supervised}}(\widetilde{I} _{\text{syn}}^i, I_{\text{real}}^i) \\
&+ \lambda_{\text{temp}} \mathcal{L}_{\text{temp}}(I_{\text{syn}}^t, I_{\text{syn}}^{t-1})) ,
\end{align}
where
\begin{align}
\mathcal{L}_{\text{GAN}}
\nonumber (I_{\text{pose}}, I_{\text{syn}}, I_{\text{real}}) 
& =  \mathbb{E}[\log D(I_{\text{pose}}, I_{\text{real}})] \\
& +  \mathbb{E} [1 - \log D(I_{\text{pose}}, I_{\text{syn}})] .
\label{equa:GAN}
\end{align}
$\mathcal{L}_{\text{supervised}}$ consists of a perceptual loss and an $L_2$ loss, which has the following form:
\begin{align}
\nonumber \mathcal{L}_{\text{supervised}}(I_{\text{syn}}, I_{\text{real}}) = & \lambda_f ||\text{VGG}(I_{\text{syn}}) - \text{VGG}(I_{\text{real}})||_1 \\
+ & \lambda_{l_2} ||I_{\text{syn}} - I_{\text{real}}||_2 .
\label{equa:supervised}
\end{align}
The perceptual loss regularizes the generated result $I_{\text{syn}}$ to be closer to the ground truth $I_{\text{real}}$ in the VGG-19 \cite{vgg19} feature space, while $L_2$ loss does similar restraints at the pixel level. \gl{ $\mathcal{L}_{\text{temp}}$ is the temporal loss term, 
and its detailed definition is given in the supplementary material. }

\textbf{Regular loss term.} Note that \yt{$\mathcal{L}_{\text{supervised}}(\widetilde{I} _{\text{syn}}^i, I_{\text{real}}^i)$ is the \emph{regular loss term} calculated from the static component from the hybrid texture. We add this term to \ytt{stabilize} the UV generator and avoid the unreasonable local optimum due to the concurrent update policy of $G_{\text{uv}}$ and $G_{\text{D}^2\text{G}}$. We visualize the effect of the regular loss in the supplementary material.}


\begin{figure*}[t]
    \centering
    \setlength{\fboxrule}{0.5pt}
    \setlength{\fboxsep}{-0.01cm}
    \begin{spacing}{1}
    \begin{tabular}{p{0.14\linewidth}<{\centering}p{0.14\linewidth}<{\centering}p{0.14\linewidth}<{\centering}p{0.14\linewidth}<{\centering}p{0.14\linewidth}<{\centering}p{0.14\linewidth}<{\centering}}

    Driving motion & \hspace{1mm}  EDN~\cite{everybody} & Ours &  
    Driving motion &  \hspace{1mm}  V2V~\cite{vid2vid} &  Ours  \\

    \includegraphics[width=1.15\linewidth]{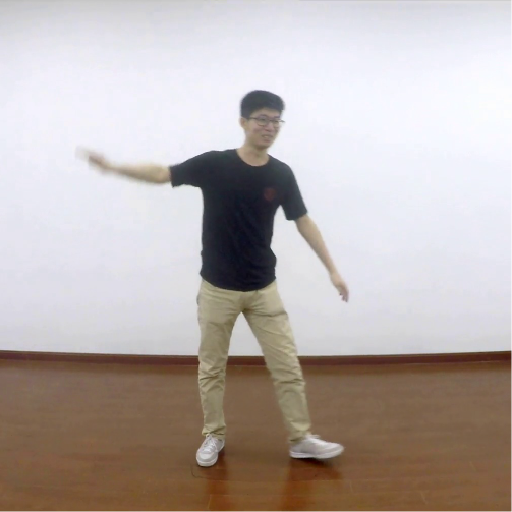} &
    \includegraphics[width=1.15\linewidth]{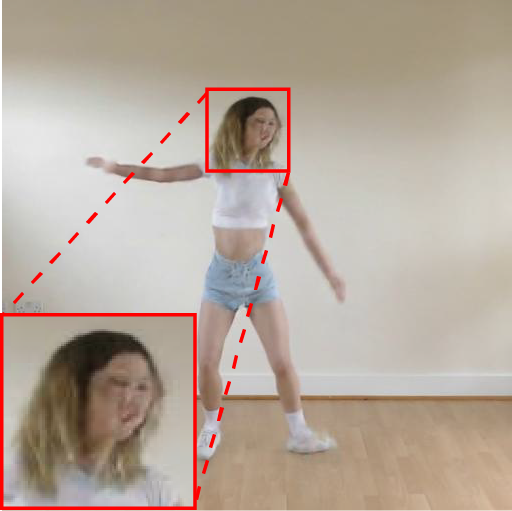}&
    \includegraphics[width=1.15\linewidth]{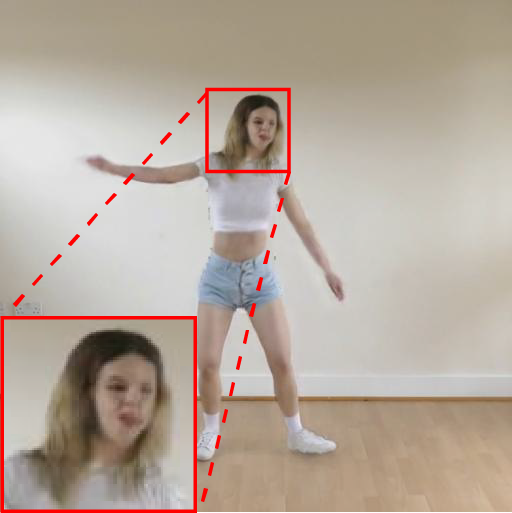} &
    \includegraphics[width=1.15\linewidth]{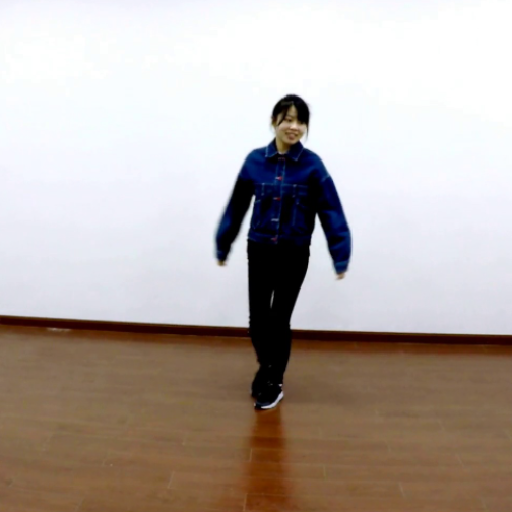} &
    \includegraphics[width=1.15\linewidth]{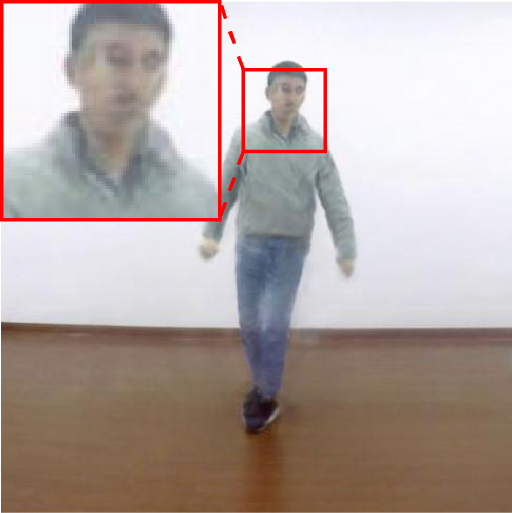}&
    \includegraphics[width=1.15\linewidth]{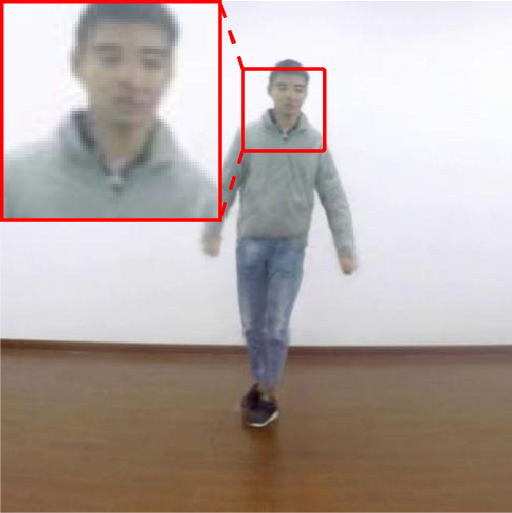}
    \\
    
    \specialrule{0em}{0pt}{-14pt} \\

    \includegraphics[width=1.15\linewidth]{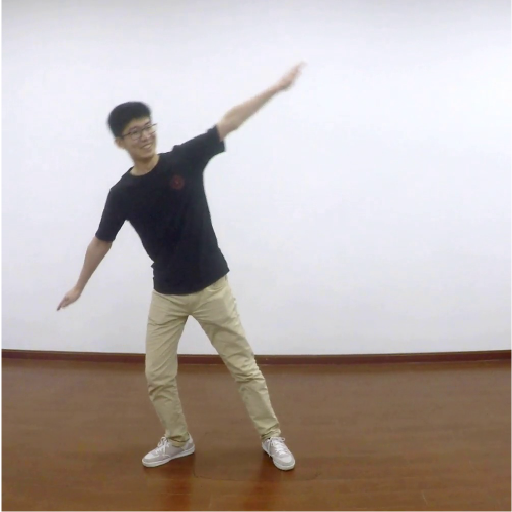} &
    \includegraphics[width=1.15\linewidth]{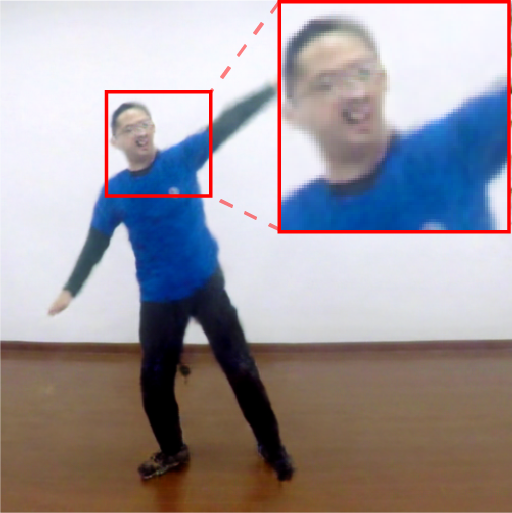}&
    \includegraphics[width=1.15\linewidth]{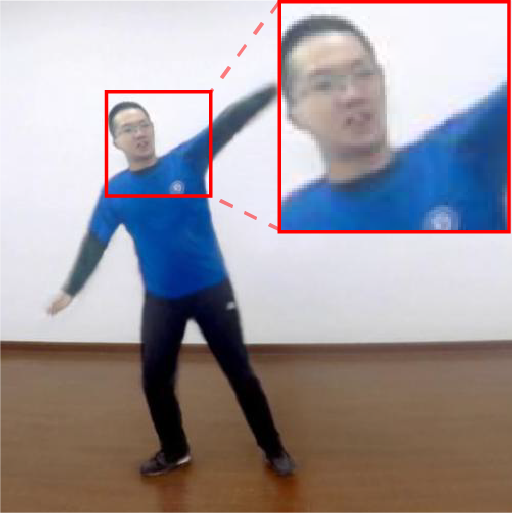} &
    \includegraphics[width=1.15\linewidth]{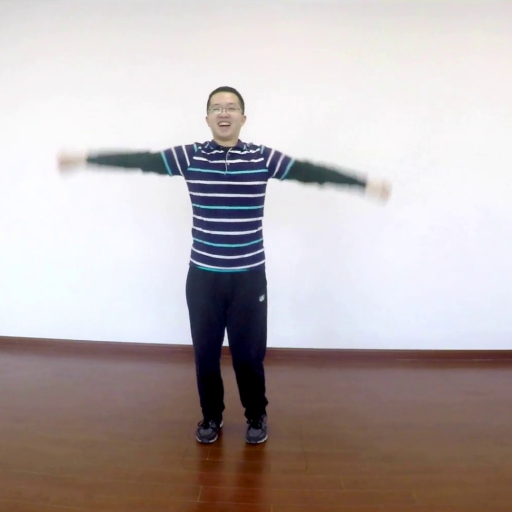} &
    \includegraphics[width=1.15\linewidth]{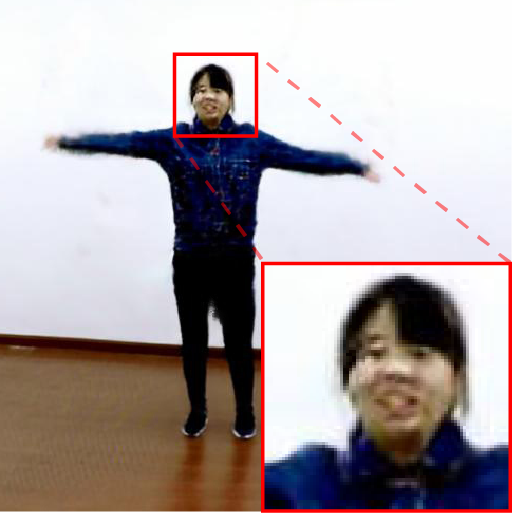}&
    \includegraphics[width=1.15\linewidth]{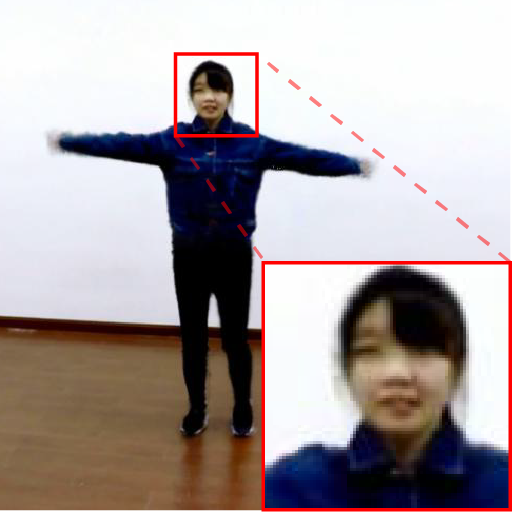}

    \end{tabular}
    \end{spacing}
    \vspace{-6mm}
    \caption{\ytt{Comparison with image-translation based approaches. We compare our approach with EDN~\cite{everybody} and V2V~\cite{vid2vid}. Our approach could produce more realistic imitating results due to the robust 3D representation, even when trained only with a short monocular video.}}
    \vspace{-3mm}
    \label{fig:compare_image_trans}
\end{figure*}

\begin{figure*}[t]
    \centering
    \setlength{\fboxrule}{0.5pt}
    \setlength{\fboxsep}{-0.01cm}
    \begin{spacing}{1}
    \begin{tabular}{p{0.14\linewidth}<{\centering}p{0.14\linewidth}<{\centering}p{0.14\linewidth}<{\centering}p{0.14\linewidth}<{\centering}p{0.14\linewidth}<{\centering}p{0.14\linewidth}<{\centering}}

    Driving motion & \hspace{1mm}  TNA~\cite{TNA} & Ours &  
    Driving motion &  \hspace{1mm}  DTL~\cite{dynamic_texture} &  Ours  \\

    \includegraphics[width=1.15\linewidth]{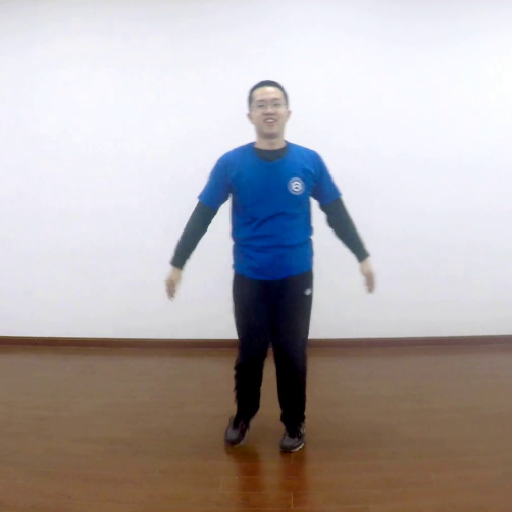} &
    \includegraphics[width=1.15\linewidth]{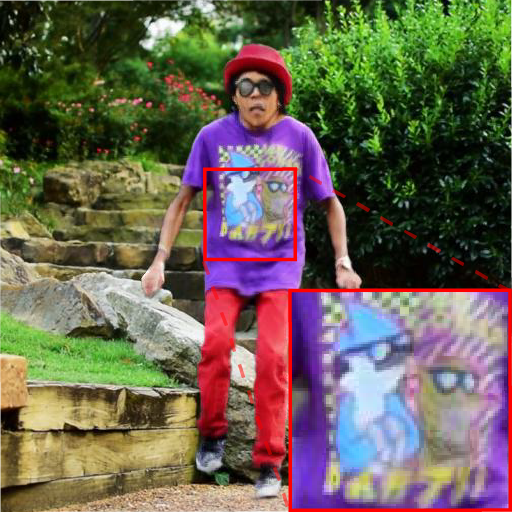}&
    \includegraphics[width=1.15\linewidth]{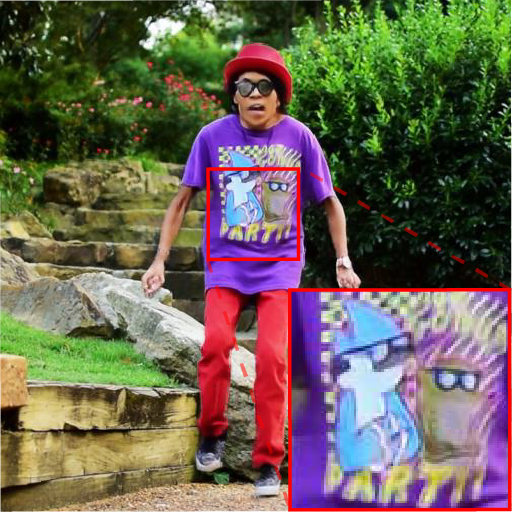} &
    \includegraphics[width=1.15\linewidth]{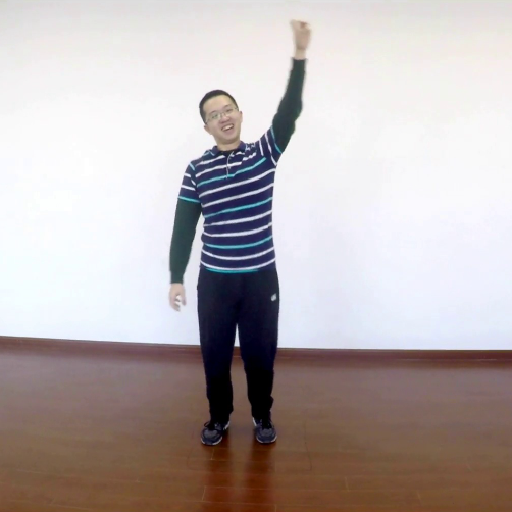} &
    \includegraphics[width=1.15\linewidth]{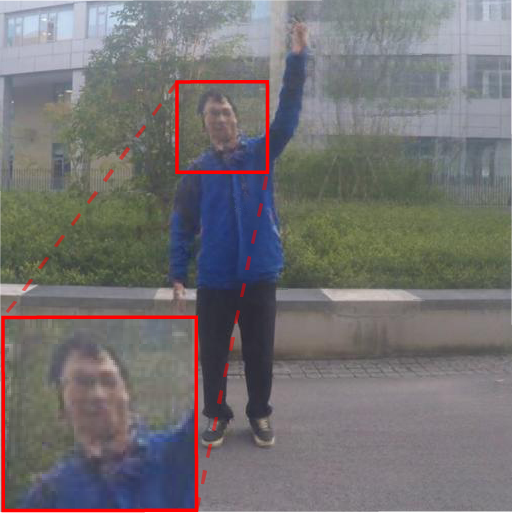} &
    \includegraphics[width=1.15\linewidth]{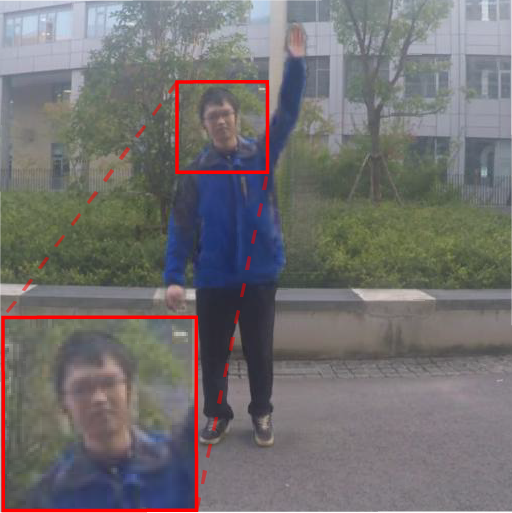}
    \\
    
    \specialrule{0em}{0pt}{-14pt} \\

    \includegraphics[width=1.15\linewidth]{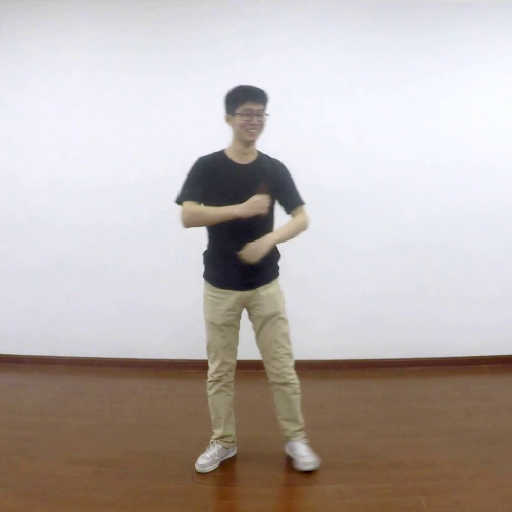} &
    \includegraphics[width=1.15\linewidth]{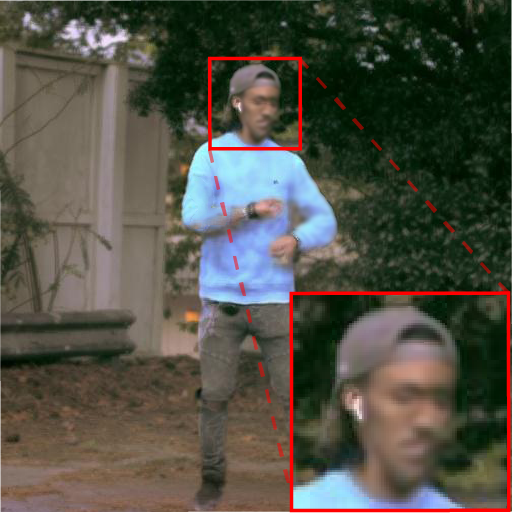}&
    \includegraphics[width=1.15\linewidth]{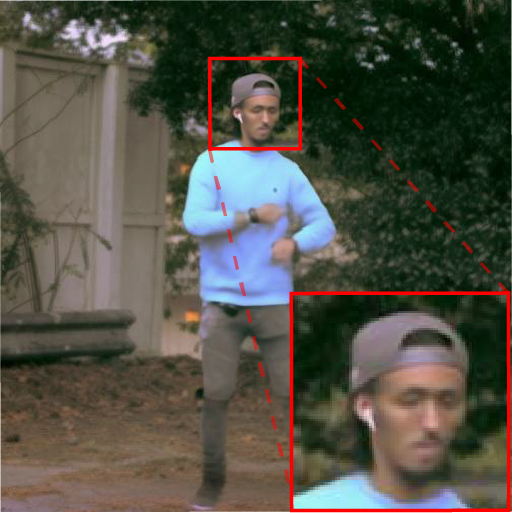} &
    \includegraphics[width=1.15\linewidth]{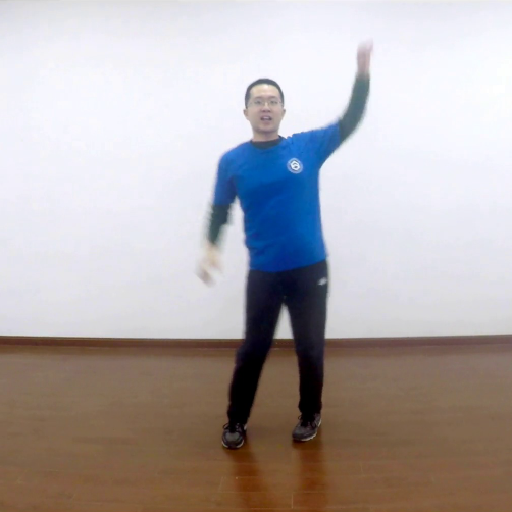} &
    \includegraphics[width=1.15\linewidth]{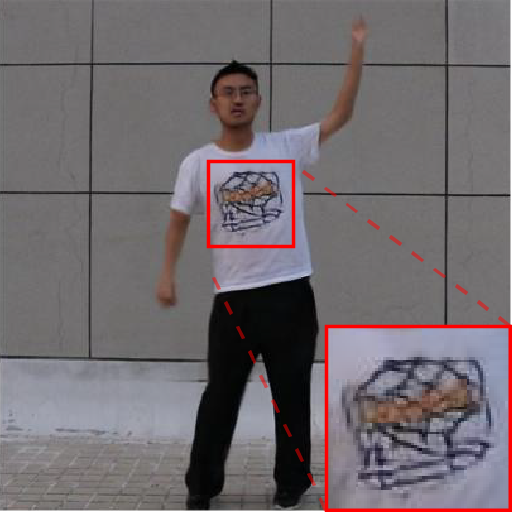} &
    \includegraphics[width=1.15\linewidth]{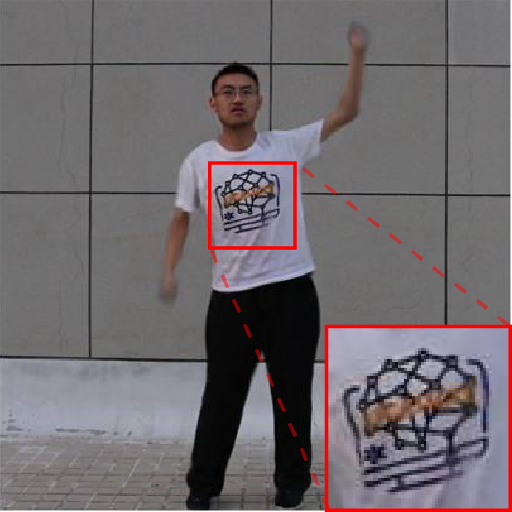}

    \end{tabular}
    \end{spacing}
    \vspace{-6mm}
    \caption{\ytt{Compared with other neural rendering baselines, our approach produces much clearer details due to the embedded image-translation components, i.e., the hybrid texture representation and D$^2$G-Net.}}
    \vspace{-6mm}
    \label{fig:compare_neural_render}
\end{figure*}

\begin{figure*}[htb]
    \centering
    \setlength{\fboxrule}{0.5pt}
    \setlength{\fboxsep}{-0.01cm}
    \begin{spacing}{1}
    \begin{tabular}{p{0.005\linewidth}<{\centering}p{0.13\linewidth}<{\centering}p{0.13\linewidth}<{\centering}p{0.13\linewidth}<{\centering}p{0.13\linewidth}<{\centering}p{0.13\linewidth}<{\centering}p{0.13\linewidth}<{\centering}}

     & Driving motion & \hspace{1mm} TNA~\cite{TNA} & Ours &  
    Driving motion &  \hspace{1mm}  DTL~\cite{dynamic_texture} &  Ours  \\
   
    \vspace{-15mm} \rotatebox{0}{$t$} &
    \includegraphics[width=1.15\linewidth]{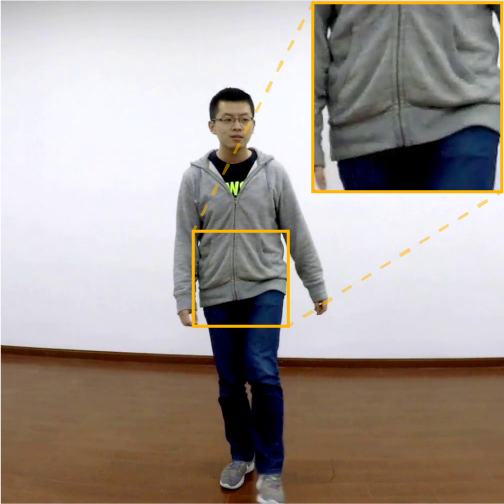} &
    \includegraphics[width=1.15\linewidth]{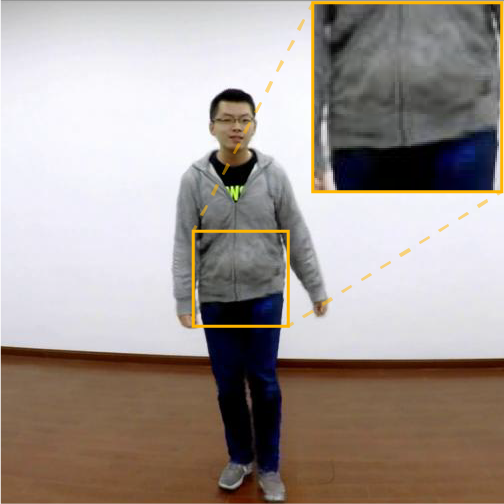} &
    \includegraphics[width=1.15\linewidth]{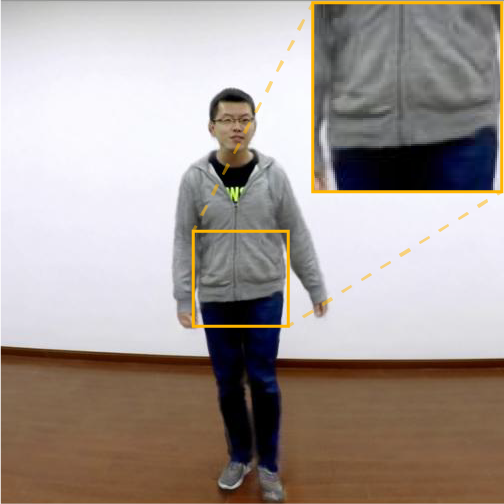} &
    \includegraphics[width=1.15\linewidth]{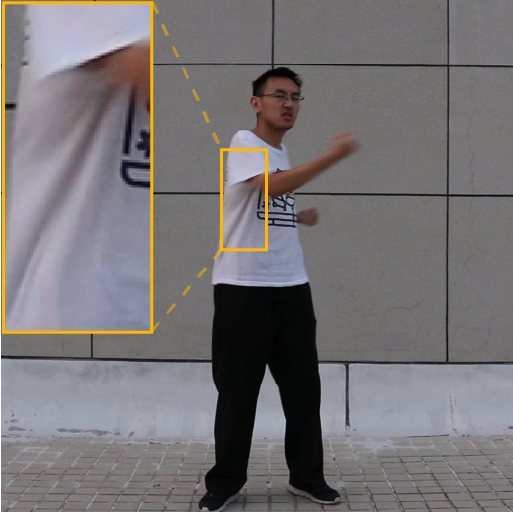} &
    \includegraphics[width=1.15\linewidth]{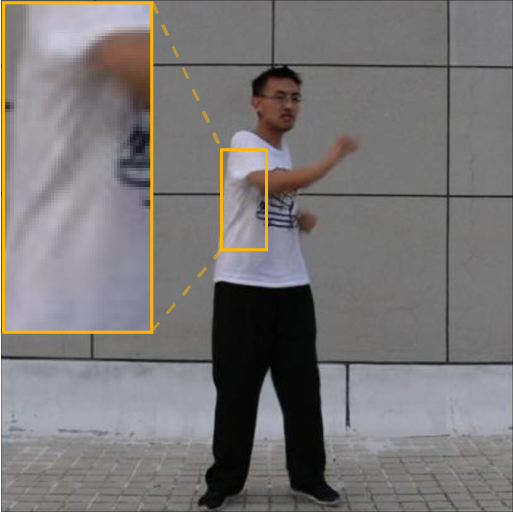} &
    \includegraphics[width=1.15\linewidth]{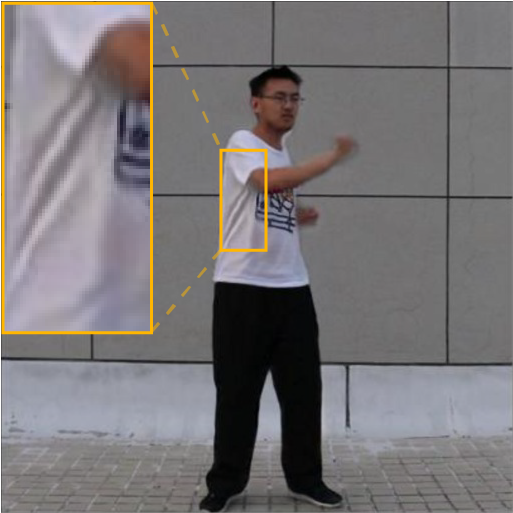} 
    \\
    \specialrule{0em}{0pt}{-14pt} \\
    
    \vspace{-15mm} \rotatebox{0}{\hspace{-2mm}$t\!+\!1$} &
    \includegraphics[width=1.15\linewidth]{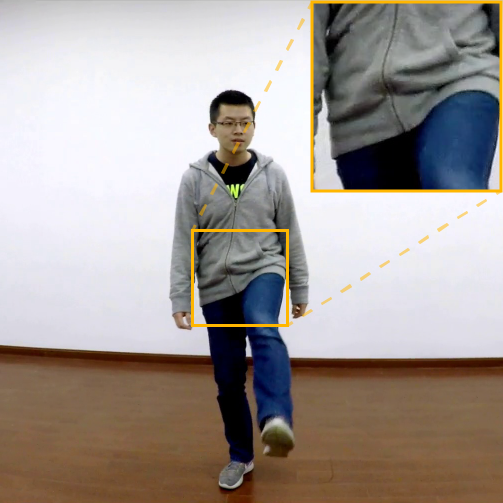} &
    \includegraphics[width=1.15\linewidth]{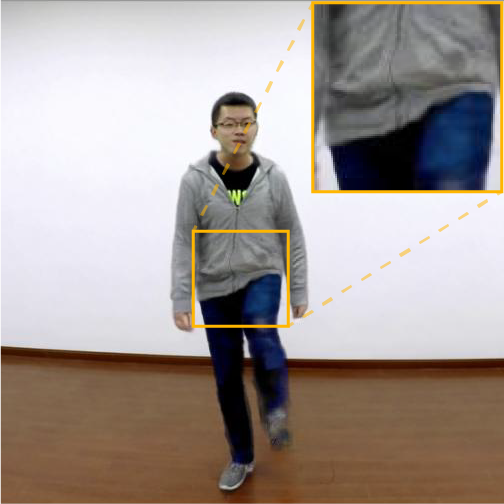} &
    \includegraphics[width=1.15\linewidth]{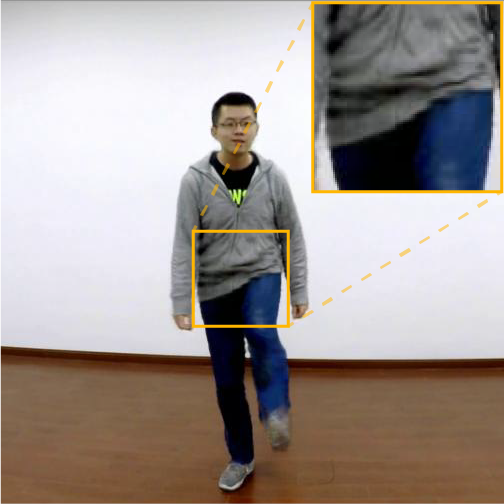} &
    \includegraphics[width=1.15\linewidth]{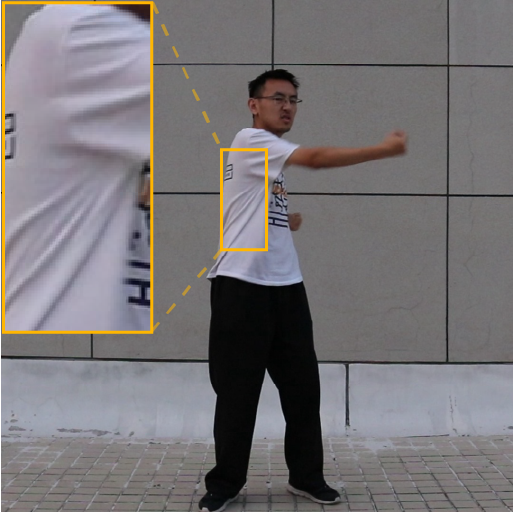} &
    \includegraphics[width=1.15\linewidth]{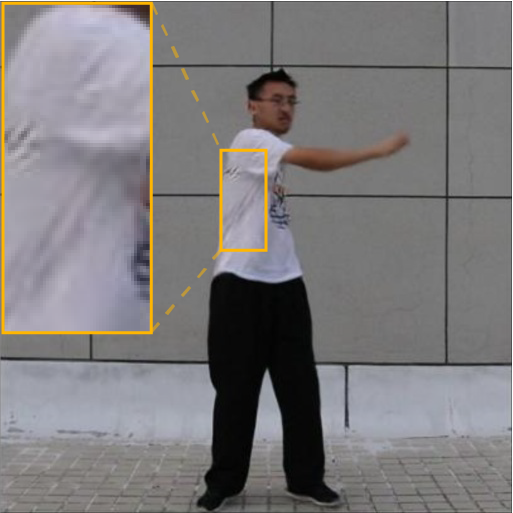} &
    \includegraphics[width=1.15\linewidth]{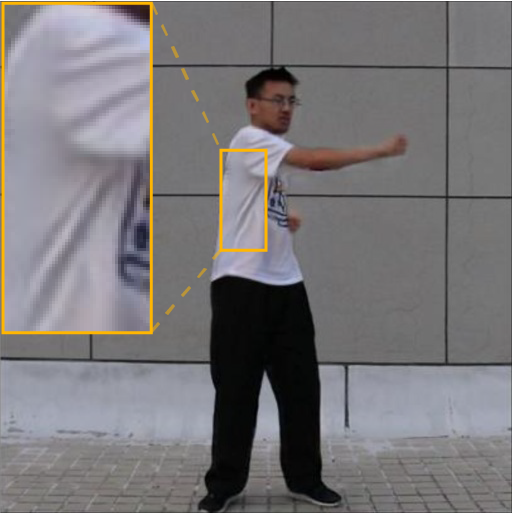} 

    \end{tabular}
    \end{spacing}
    \vspace{-6mm}
    \caption{Comparison of \ytt{pose-aware} dynamic details. The two rows correspond to consecutive frames of $t$ and $t+1$, respectively. Compared with TNA~\cite{TNA} and DTL~\cite{dynamic_texture}, our method better delineates details for different poses. We show the results of self-transfer for the reference of ground truth. More pose transfer results will be exhibited in the supplementary material.    }
    \vspace{-3mm}
    \label{fig:compare_dynamic}
\end{figure*}

\vspace{-3mm}
\section{Experiments}
\begin{figure}[t]
    \centering
    \setlength{\fboxrule}{0.5pt}
    \setlength{\fboxsep}{-0.01cm}
    \begin{spacing}{1}
    \begin{tabular}{p{0.28\linewidth}<{\centering}p{0.28\linewidth}<{\centering}p{0.28\linewidth}<{\centering}}
    
   Background 1 &  \hspace{3mm}  Background 2 &  Background 3  \\
    \includegraphics[width=1.15\linewidth]{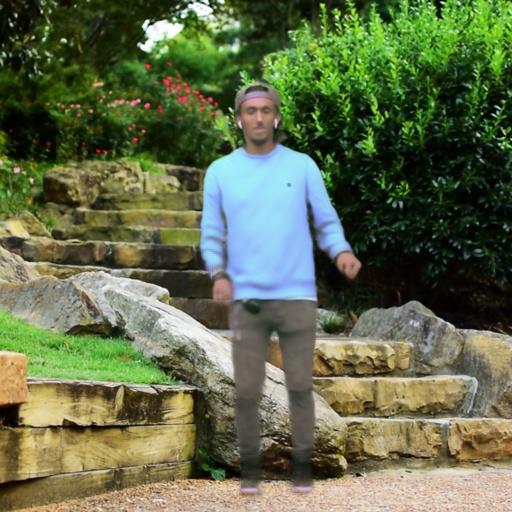} & 
    \includegraphics[width=1.15\linewidth]{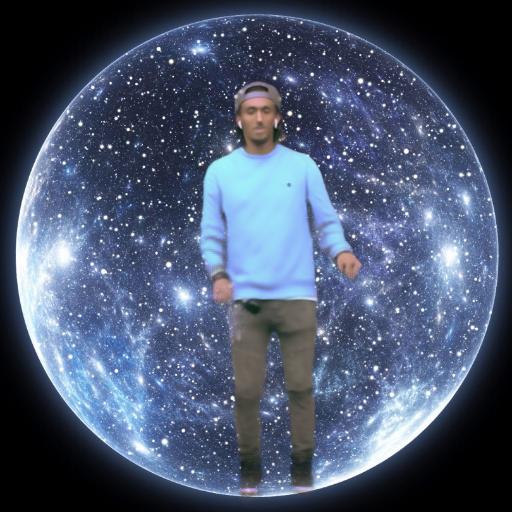} & 
    \includegraphics[width=1.15\linewidth]{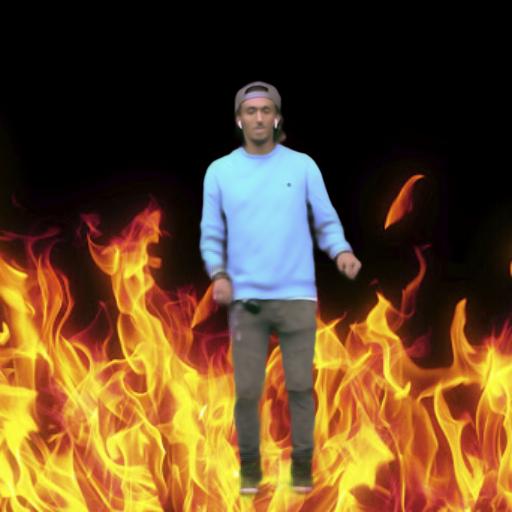} 

    \end{tabular}
    \end{spacing}
    \vspace{-6mm}
    \caption{Examples of the same generated person with different backgrounds. Our approach is able to generate videos with arbitrary novel backgrounds by replacing the refined background image.
    }
    \vspace{-6mm}
    \label{fig:background}
\end{figure}

\yt{We perform sufficient comparisons to demonstrate the advantage of our approach, which preserves dynamic texture details while only relies on a single accessible monocular video for training.
We compare our approach with the state-of-the-art methods under the same data setting. We also verify the importance of each key component in our framework, including the image translation components, pose label condition in the D$^2$G-Net, the regular loss term and the temporal loss.}
\ytt{Moreover, benefiting from the learned precise mask of human foreground,
our approach is capable of replacing the background in the synthesized video frames. One example is presented in Fig.~\ref{fig:background}.
}

\vspace{-3mm}
\subsection{Datasets and Metrics}
We conduct massive experiments on the dataset consisting of selected videos from iPER dataset~\cite{LWGAN}, several online videos and one own video, each of which lasts about 3 minutes with 25 FPS (2$\sim$4k valid frames). To ensure the data diversity, there are multiple clothes and actions. 
\ytt{The performance of the generative model is evaluated under two conditions depending on whether the source and target characters are the same (self-transfer) or different (cross-transfer). For the former, the Structural Similarity Index Measure (SSIM)~\cite{SSIM} and Peak Signal-to-Noise Ratio (PSNR) are used to indicate the quality of generated results; for the latter, we use the Fréchet Inception Distance (FID) \cite{FID} with features extracted by InceptionV3~\cite{szegedy2016rethinking} network similar with~\cite{TNA}, because there are no ground truth.}

However, the above metrics can only reflect the quality of generated results from the holistic perspective. To better measure the robustness of the generative model, we propose a new metric which focuses on poses with a large deviation from the training sample (denoted as ``\emph{challenging pose}''). 
To be more specific, for each pose $i$ in the validation set, we compute its nearest neighbor distance $d_i$ with training samples. Here the distance between two poses is characterized by Euler distance of 2D keypoints.
Assuming the poses with larger value of $d_i$ are more challenging, $M$ validation samples ($M=10$ in our experiments) with largest $d_i$ are selected to form the challenging poses. We evaluate the model performance under these poses with SSIM and PSNR, which is denoted as ``R-SSIM''(Robust SSIM) and ``R-PSNR''(Robust-PSNR) for clarity.

\ytt{In addition to these metrics calculated on single frames, we introduce the metric \textit{temporal error} $E_{\text{temp}}$ from~\cite{temporal_loss} to evaluate the temporal coherence of generated image sequences. We elaborate the definition of $E_{\text{temp}}$ in the supplementary material.
}

\vspace{-3mm}
\subsection{Comparison with State of the Arts (SOTAs)}
We compare our method against existing state-of-the-art methods Vid2vid \cite{vid2vid} (\ytt{V2V}), Everybody Dance Now \cite{everybody} (EDN) and Liquid Warping GAN \cite{LWGAN} (LWG), using official implementations. We also compare with existing neural rendering methods Texture Neural Avatar~\cite{TNA} (TNA) and Dynamic Texture Learning~\cite{dynamic_texture} (DTL) based on our re-implementation with the same experimental settings as original, since no source code is available. 
Note that for TNA, we add our refined background for fair comparison.
As for DTL, we use the off-the-shelf UV parameterization (DensePose~\cite{densepose}) \ytt{for frame unwrapping} since the accurate 3D reconstruction which relies on multi-view capture devices is not available. 
We show those comparison results with SOTA in Table~\ref{table:compare}. 
Moreover, since LWG model is not limited to a particular person, we only list the results for reference, but it does not participate in comparison.

\begin{table}[h]
\setlength\tabcolsep{1pt}
\begin{center}
\vspace{-3mm}
\caption{Quantitative comparison with SOTA methods. Our approach can synthesize more reasonable results balancing image quality and robustness. Ours also outperforms the others in temporal coherence. The numbers in italics are for reference only, not for comparison.}
\label{table:compare}
\vspace{-6mm}
\resizebox{\linewidth}{!}{
\begin{tabular}{lcccccc}
\hline
Method & \small{LWG~\cite{LWGAN}} & \small{EDN~\cite{everybody}}  &  \small{V2V~\cite{vid2vid}} & \small{TNA~\cite{TNA}} &\small{DTL~\cite{dynamic_texture}} & \small{Ours} \\ \hline
\small{SSIM}    $\uparrow$           & \textit{0.821}             & 0.911             & 0.924                & 0.925                      & 0.924             & \ytt{\textbf{0.933}}  \\ 
\small{R-SSIM}  $\uparrow$   &   \textit{0.823}            & 0.906           & 0.915                & 0.918                      & 0.917            & \ytt{\textbf{0.930}} \\ 
\small{PSNR} $\uparrow$              & \textit{32.58}             & 37.10              & 36.60                 & 36.80                       & 37.2              & \ytt{\textbf{37.44}}  \\ 
\small{R-PSNR}  $\uparrow$   & \textit{32.44}               & 36.41           & 35.91              & 36.62                        & 36.50            & \ytt{\textbf{36.99}}  \\ 
\small{Temporal} $\downarrow$        & \textit{0.78}              & 0.61              & 0.48                 & 0.46                       & 0.51              & \ytt{\textbf{0.41}}  \\ 
\small{FID}      $\downarrow$        & \textit{71.20}             & 58.86             & 57.04               & 55.68                      & 56.74             & \ytt{\textbf{53.49}}  \\ \hline
\end{tabular}
}
\end{center}
\end{table}
\vspace{-3mm}

{\bf Comparison with direct image translation approaches.}
\ytt{The direct image-translation based approaches~\cite{everybody, vid2vid} suffer from a lack of explicit 3D representation, which leads to poor generalization ability especially on short training videos, as illustrated in Fig.~\ref{fig:compare_image_trans}. There is also a more significant drop of quantitative score for the ``\emph{challenging pose}'', as shown in  ``R-SSIM``and``R-PSNR`` in  Table~\ref{table:compare}. 
\ytFinal{We further demonstrate the dependence of such methods on the training data length by evaluating their performance under different numbers of training frames in the supplementary material.}
}

{\bf Comparison with neural rendering approaches.}
Compared with other neural rendering approaches (TNA~\cite{TNA} and DTL~\cite{dynamic_texture}), our method produces the results \yt{with richer details and higher fidelity under the same data configuration, \ytt{as illustrated in  Fig.~\ref{fig:compare_neural_render}. We also visualize a pose-aware dynamic details comparison in Fig.~\ref{fig:compare_dynamic}.} 
Although TNA can still produce reasonable results when the training data is limited to monocular video, its static texture map goes against to the dynamic details generation under different poses. By contrast our approach achieves better performance by interpreting the high-dimensional texture feature to details dynamically.
\ytt{As for DTL, the dynamic characteristics are learned explicitly under the supervision of back projected frames according to a finely reconstructed 3D model. Due to the inevitable reconstruction errors with monocular training video, the performance degrades significantly. In contrast, our approach models the dynamic details with implicit image-translation components, which can be learned automatically with only the supervision on final rendering images. Such novel representation and end-to-end training framework make our approach outperform DTL.}
}


Overall, our method synthesizes high quality results with rich high-frequency details, and is more robust to the challenging inputs compared with existing SOTA methods.





{\bf User study.}
We conduct a user study to measure the human perceptual quality for pose transfer results
\ytt{and our results are considered more realistic than baseline approaches. These comparisons are given in the supplementary material.} 

\vspace{-4mm}
\subsection{Ablation Study}
\begin{figure*}[thb]
    \centering
    \setlength{\fboxrule}{0.5pt}
    \setlength{\fboxsep}{-0.01cm}
    \begin{spacing}{1}
    \begin{tabular}{p{0.14\linewidth}<{\centering}p{0.14\linewidth}<{\centering}p{0.14\linewidth}<{\centering}p{0.14\linewidth}<{\centering}p{0.14\linewidth}<{\centering}p{0.14\linewidth}<{\centering}}
    
   Driving motion & w/o D$^2$G & w/o HybridTex  &  w/o Pose cond & MaskTex & Ours  \\
    \includegraphics[width=1.15\linewidth]{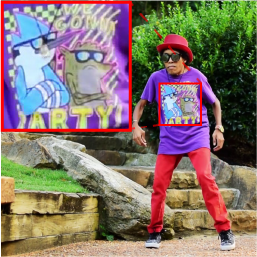} &
    \includegraphics[width=1.15\linewidth]{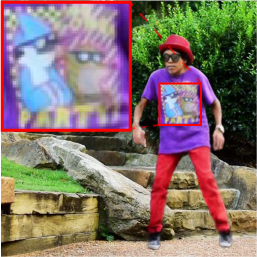} &
    \includegraphics[width=1.15\linewidth]{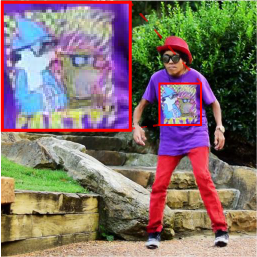} &
    \includegraphics[width=1.15\linewidth]{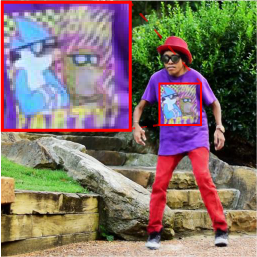} &
    \includegraphics[width=1.15\linewidth]{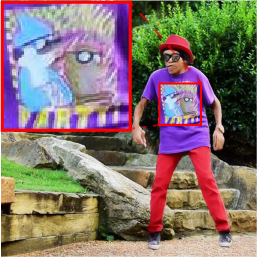} &
    \includegraphics[width=1.15\linewidth]{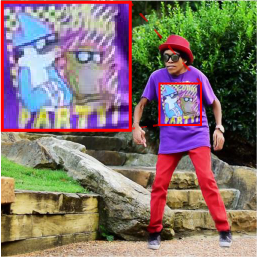} 
    \\
    \specialrule{0em}{0pt}{-14pt} \\
    \includegraphics[width=1.15\linewidth]{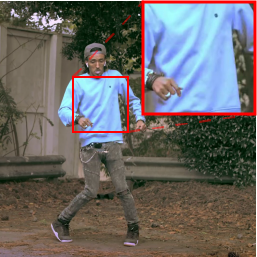} &
    \includegraphics[width=1.15\linewidth]{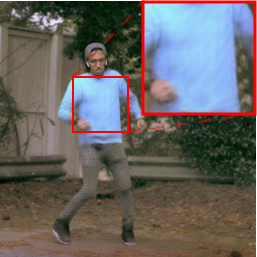} &
    \includegraphics[width=1.15\linewidth]{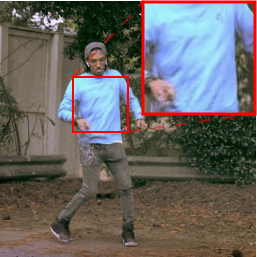} &
    \includegraphics[width=1.15\linewidth]{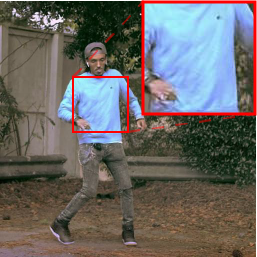} &
    \includegraphics[width=1.15\linewidth]{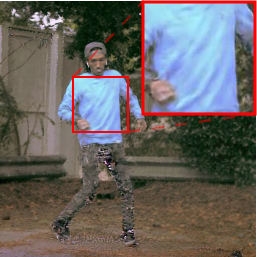} &
    \includegraphics[width=1.15\linewidth]{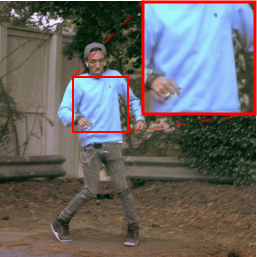} 
   
    \end{tabular}
    \end{spacing}
    \vspace{-6mm}
    \caption{\ytt{We visualize the effect of key components in our approach. Compared with other variants, our approach generates clearer high-frequency details (1st row) and pose-varying wrinkles (2nd row).}}
    \vspace{-6mm}
    \label{fig:ablation}
\end{figure*}

\ytt{We evaluate the role of each component in our pipeline via convictive ablation studies. Specifically, we investigate how the image translation components and pose label condition work in the generation of pose-aware high-frequency details, as reported in Fig.~\ref{fig:ablation} and Table~\ref{table:image_translation}. We also demonstrate the importance of regular loss and temporal loss in the supplementary material.}

{\bf Effect of image translation component.}
\label{subsec:albation_image_translation}
\ytt{The ability of our approach to characterize high-frequency signals is largely benefited from the image-translation component embedded in the neural rendering framework, i.e. the D$^2$G-Net and the novel hybrid texture representation. We have performed experiments without the D$^2$G network (``w/o D$^2$G``) and with common RGB texture representation (``w/o HybridTex``) to demonstrate their effect respectively, as illustrated in the 2nd and 3rd columns of Fig.~\ref{fig:ablation}.
Without the D$^2$G-Net, the pipeline actually degrades to the static texture and the dynamic details cannot be well exhibited. 
When the hybrid texture is removed, the rendered high-frequency details become diminished and unrealistic, which proves the significance of the high-level appearance representation.}
\ytt{We further verify the effect of the hybrid texture by masking out the extra channels of the learned texture in our whole pipeline, i.e., the extra channels are set to zero (``MaskTex''). The result is reported in the 5th column of Fig.~\ref{fig:ablation} and it can be seen that the high-frequency signals are weakened due to the absence of the hybrid representation.}

{\bf Relevance of pose label condition.}
\yt{To characterize the dynamic details of human surface from hybrid texture under the varying poses, we condition the  D$^2$G-Net with current pose labels. Note that although there exists implicit pose information in the human foreground of texture feature, the varying feature during training process would disturb the converge of the translation network. \ytt{We report the result without pose label condition (denoted as ``w/o Pose cond``) in Fig.~\ref{fig:ablation}.}}




\begin{table}[t]
\begin{center}
\vspace{-3mm}
\caption{The ablation study for the effect of the image translation component. Our full pipeline scores best than other variants, which demonstrates the effect of image translation components and pose label condition.}
\label{table:image_translation}
\vspace{-6mm}
\resizebox{\linewidth}{!}{
\begin{tabular}{lcccc}
\hline
\small{Method}  & \small{\ytt{w/o D$^2$G}}  & \small{\ytt{w/o HybridTex}} & \small{\ytt{w/o Pose cond}} & \small{\ytt{Ours}} \\ \hline
\small{SSIM} $\uparrow$   & \ytt{0.913}              & \ytt{0.921}              & \ytt{0.918}       & \ytt{\textbf{0.933}}  \\ 
\small{PSNR} $\uparrow$    & \ytt{36.35}              & \ytt{36.87}              & \ytt{36.62}       & \ytt{\textbf{37.44}}   \\ \hline
\end{tabular}
}
\end{center}
\vspace{-6mm}
\end{table}

\vspace{-4mm}
\section{Limitations and Discussion}
Although our approach is able to generate coherent videos with high fidelity, there are still several limitations. First, we have to train a new model for each specific person, which greatly limits the applicable scenarios of our method. 
\gl{Second,} since our pipeline is based on existing pose tracking approaches, tracking errors would also affect the quality of generated results. \ytt{This could be alleviated by the accuracy improvement of pose estimation. 
\gl{Third,} current approach is limited to the training video with a static background. 
It will further improve the flexibility by extending applicable scenes to having dynamic background in the future work.}


\vspace{-3mm}
\section{Conclusion}
\yt{
We have proposed a new approach for pose transfer on the human video synthesis task. 
By embedding the learnable hybrid texture into the neural rendering pipeline with the dynamic details generation mechanism, our approach is able to generate high-fidelity human video frames in an end-to-end manner.
In this means, our approach avoids the dependency for abundant training data and serves as a more accessible approach. 
Furthermore, 
\ytt{the pre-training strategy and additional spatial-temporal losses together help regularize the network.}
\ytt{Overall, this approach outperforms SOTAs in both robustness and fidelity with training data from short monocular videos. We would provide the source code of PyTorch and Jittor~\cite{hu2020jittor} in the future. Jittor is a fully just-in-time (JIT) compiled deep learning framework.}}

\vspace{-3mm}
\section*{Acknowledgment}
This work was supported by the Beijing Municipal Natural Science Foundation for Distinguished Young Scholars (No. JQ21013), the National Natural Science Foundation of China (No. 62061136007 and No. 61872440), Royal Society Newton Advanced Fellowship (No. NAF\verb|\|R2\verb|\|192151), Tencent AI Lab Rhino-Bird Focused Research Program and the Youth Innovation Promotion Association CAS.

\vspace{-3mm}
{\small
    \bibliographystyle{IEEEtran}
    \bibliography{IEEEabrv,egbib}

\begin{thebibliography}{10}
\providecommand{\url}[1]{#1}
\csname url@samestyle\endcsname
\providecommand{\newblock}{\relax}
\providecommand{\bibinfo}[2]{#2}
\providecommand{\BIBentrySTDinterwordspacing}{\spaceskip=0pt\relax}
\providecommand{\BIBentryALTinterwordstretchfactor}{4}
\providecommand{\BIBentryALTinterwordspacing}{\spaceskip=\fontdimen2\font plus
\BIBentryALTinterwordstretchfactor\fontdimen3\font minus
  \fontdimen4\font\relax}
\providecommand{\BIBforeignlanguage}[2]{{%
\expandafter\ifx\csname l@#1\endcsname\relax
\typeout{** WARNING: IEEEtran.bst: No hyphenation pattern has been}%
\typeout{** loaded for the language `#1'. Using the pattern for}%
\typeout{** the default language instead.}%
\else
\language=\csname l@#1\endcsname
\fi
#2}}
\providecommand{\BIBdecl}{\relax}
\BIBdecl

\bibitem{everybody}
C.~Chan, S.~Ginosar, T.~Zhou, and A.~A. Efros, ``Everybody dance now,'' in
  \emph{International Conference on Computer Vision}, 2019, pp. 5932--5941.

\bibitem{vid2vid}
T.-C. Wang, M.-Y. Liu, J.-Y. Zhu, G.~Liu, A.~Tao, J.~Kautz, and B.~Catanzaro,
  ``Video-to-video synthesis,'' in \emph{Advances in Neural Information
  Processing Systems}, 2018, p. 1152–1164.

\bibitem{Dense_Intrinsic_Appearance_Flow}
Y.~Li, C.~Huang, and C.~C. Loy, ``Dense intrinsic appearance flow for human
  pose transfer,'' in \emph{IEEE Conference on Computer Vision and Pattern
  Recognition}, 2019, pp. 3688--3697.

\bibitem{Deep_Motion_Transfer}
A.~Siarohin, S.~Lathuilière, S.~Tulyakov, E.~Ricci, and N.~Sebe, ``Animating
  arbitrary objects via deep motion transfer,'' in \emph{IEEE Conference on
  Computer Vision and Pattern Recognition}, 2019, pp. 2372--2381.

\bibitem{PG}
L.~Ma, X.~Jia, Q.~Sun, B.~Schiele, T.~Tuytelaars, and L.~Van~Gool, ``Pose
  guided person image generation,'' in \emph{Advances in Neural Information
  Processing Systems}, 2017, pp. 405--415.

\bibitem{First_Order_Motion_Model}
A.~Siarohin, S.~Lathuilière, S.~Tulyakov, E.~Ricci, and N.~Sebe, ``First order
  motion model for image animation,'' in \emph{Advances in Neural Information
  Processing Systems}, 2019, p. 7137–7147.

\bibitem{TNA}
A.~Shysheya, E.~Zakharov, K.-A. Aliev, R.~Bashirov, E.~Burkov, K.~Iskakov,
  A.~Ivakhnenko, Y.~Malkov, I.~Pasechnik, D.~Ulyanov, A.~Vakhitov, and V.~S.
  Lempitsky, ``Textured neural avatars,'' 2019, pp. 2382--2392.

\bibitem{densepose_transfer}
N.~Neverova, R.~A. G{\"u}ler, and I.~Kokkinos, ``Dense pose transfer,'' in
  \emph{European Conference on Computer Vision}, 2018, pp. 128--143.

\bibitem{dynamic_texture}
L.~Liu, W.~Xu, M.~Habermann, M.~Zollh{\"o}fer, F.~Bernard, H.~Kim, W.~Wang, and
  C.~Theobalt, ``Neural human video rendering by learning dynamic textures and
  rendering-to-video translation,'' \emph{IEEE Transactions on Visualization
  and Computer Graphics}, vol.~27, pp. 1--1, 05 2021.

\bibitem{temporal_loss}
H.~{Huang}, H.~{Wang}, W.~{Luo}, L.~{Ma}, W.~{Jiang}, X.~{Zhu}, Z.~{Li}, and
  W.~{Liu}, ``Real-time neural style transfer for videos,'' in \emph{IEEE
  Conference on Computer Vision and Pattern Recognition}, 2017, pp. 7044--7052.

\bibitem{video_rewrite}
C.~Bregler, M.~Covell, and M.~Slaney, ``Video rewrite: driving visual speech
  with audio,'' in \emph{Proceedings of the annual conference on Computer
  graphics and interactive techniques}, 1997, p. 353–360.

\bibitem{cGAN}
M.~Mirza and S.~Osindero, ``Conditional generative adversarial nets,''
  \emph{ArXiv}, vol. abs/1411.1784, 2014.

\bibitem{cycleGAN2017}
J.-Y. Zhu, T.~Park, P.~Isola, and A.~A. Efros, ``Unpaired image-to-image
  translation using cycle-consistent adversarial networks,'' in
  \emph{International Conference on Computer Vision}, 2017, pp. 2242--2251.

\bibitem{discoGAN}
T.~Kim, M.~Cha, H.~Kim, J.~K. Lee, and J.~Kim, ``Learning to discover
  cross-domain relations with generative adversarial networks,'' in
  \emph{International Conference on Machine Learning}, 2017, p. 1857–1865.

\bibitem{thiesDeferred2019}
J.~Thies, M.~Zollh{\"o}fer, and M.~Nie{\ss}ner, ``Deferred neural rendering,''
  \emph{ACM Transactions on Graphics (TOG)}, vol.~38, pp. 1 -- 12, 2019.

\bibitem{lombardiDeep2018}
S.~Lombardi, J.~M. Saragih, T.~Simon, and Y.~Sheikh, ``Deep appearance models
  for face rendering,'' \emph{ACM Transactions on Graphics (TOG)}, vol.~37, pp.
  1 -- 13, 2018.

\bibitem{liuNeural2019}
L.~Liu, W.~Xu, M.~Zollhoefer, H.~Kim, F.~Bernard, M.~Habermann, W.~Wang, and
  C.~Theobalt, ``Neural rendering and reenactment of human actor videos,''
  \emph{ACM Transactions on Graphics (TOG)}, vol.~38, pp. 1 -- 14, 2019.

\bibitem{pix2pix}
P.~Isola, J.-Y. Zhu, T.~Zhou, and A.~A. Efros, ``Image-to-image translation
  with conditional adversarial networks,'' 2017, pp. 5967--5976.

\bibitem{unet}
O.~Ronneberger, P.~Fischer, and T.~Brox, ``U-net: Convolutional networks for
  biomedical image segmentation,'' in \emph{International Conference on Medical
  Image Computing and Computer Assisted Intervention}, 2015, pp. 234--241.

\bibitem{pix2pixHD}
T.-C. Wang, M.-Y. Liu, J.-Y. Zhu, A.~Tao, J.~Kautz, and B.~Catanzaro,
  ``High-resolution image synthesis and semantic manipulation with conditional
  {GANs},'' 2018, pp. 8798--8807.

\bibitem{LWGAN}
W.~Liu, Z.~Piao, J.~Min, W.~Luo, L.~Ma, and S.~Gao, ``Liquid warping {GAN}: A
  unified framework for human motion imitation, appearance transfer and novel
  view synthesis,'' in \emph{International Conference on Computer Vision},
  2019, pp. 5903--5912.

\bibitem{posewarp}
G.~Balakrishnan, A.~Zhao, A.~V. Dalca, F.~Durand, and J.~V. Guttag,
  ``Synthesizing images of humans in unseen poses,'' 2018, pp. 8340--8348.

\bibitem{Siarohin2018DeformableGF}
A.~Siarohin, E.~Sangineto, S.~Lathuili{\`e}re, and N.~Sebe, ``Deformable {GANs}
  for pose-based human image generation,'' 2018, pp. 3408--3416.

\bibitem{Esser2018AVU}
P.~Esser, E.~Sutter, and B.~Ommer, ``A variational {U-Net} for conditional
  appearance and shape generation,'' 2018, pp. 8857--8866.

\bibitem{Ma2018DisentangledPI}
L.~Ma, Q.~Sun, S.~Georgoulis, L.~V. Gool, B.~Schiele, and M.~Fritz,
  ``Disentangled person image generation,'' \emph{IEEE Conference on Computer
  Vision and Pattern Recognition}, pp. 99--108, 2018.

\bibitem{videoGAN}
C.~Vondrick, H.~Pirsiavash, and A.~Torralba, ``Generating videos with scene
  dynamics,'' in \emph{Advances in Neural Information Processing Systems},
  2016, p. 1857–1865.

\bibitem{mocoGAN}
S.~Tulyakov, M.-Y. Liu, X.~Yang, and J.~Kautz, ``{MoCoGAN}: Decomposing motion
  and content for video generation,'' \emph{IEEE Conference on Computer Vision
  and Pattern Recognition}, pp. 1526--1535, 2018.

\bibitem{openpose}
Z.~{Cao}, G.~{Hidalgo Martinez}, T.~{Simon}, S.~{Wei}, and Y.~A. {Sheikh},
  ``{OpenPose}: Realtime multi-person {2D} pose estimation using part affinity
  fields,'' \emph{IEEE Transactions on Pattern Analysis and Machine
  Intelligence}, pp. 1302--1310, 2019.

\bibitem{total_capture}
D.~Xiang, H.~Joo, and Y.~Sheikh, ``Monocular total capture: Posing face, body,
  and hands in the wild,'' in \emph{IEEE Conference on Computer Vision and
  Pattern Recognition}, 2019, pp. 10\,957--10\,966.

\bibitem{human_motion_tranfer_3d}
Y.~Sun, Q.~Fu, Y.~Jiang, Z.~Liu, Y.~Lai, H.~Fu, and L.~Gao, ``Human motion
  transfer with {3D} constraints and detail enhancement,'' \emph{ArXiv}, vol.
  abs/2003.13510, 2020.

\bibitem{densepose}
R.~A. G{\"u}ler, N.~Neverova, and I.~Kokkinos, ``{DensePose}: Dense human pose
  estimation in the wild,'' in \emph{IEEE Conference on Computer Vision and
  Pattern Recognition}, 2018, pp. 7297--7306.

\bibitem{deepfillv2}
J.~Yu, Z.~Lin, J.~Yang, X.~Shen, X.~Lu, and T.~S. Huang, ``Free-form image
  inpainting with gated convolution,'' in \emph{International Conference on
  Computer Vision}, 2019, pp. 4470--4479.

\bibitem{vgg19}
K.~Simonyan and A.~Zisserman, ``Very deep convolutional networks for
  large-scale image recognition,'' \emph{ArXiv}, vol. abs/1409.1556, 2015.

\bibitem{SSIM}
Z.~Wang, A.~C. Bovik, H.~R. Sheikh, and E.~P. Simoncelli, ``Image quality
  assessment: from error visibility to structural similarity,'' \emph{IEEE
  Transactions on Image Processing}, vol.~13, pp. 600--612, 2004.

\bibitem{FID}
M.~Heusel, H.~Ramsauer, T.~Unterthiner, B.~Nessler, and S.~Hochreiter, ``Gans
  trained by a two time-scale update rule converge to a local nash
  equilibrium,'' in \emph{Advances in Neural Information Processing Systems},
  2017, p. 6629–6640.

\bibitem{szegedy2016rethinking}
C.~Szegedy, V.~Vanhoucke, S.~Ioffe, J.~Shlens, and Z.~Wojna, ``Rethinking the
  inception architecture for computer vision,'' in \emph{IEEE Conference on
  Computer Vision and Pattern Recognition}, 2016, pp. 2818--2826.

\bibitem{hu2020jittor}
S.-M. Hu, D.~Liang, G.-Y. Yang, G.-W. Yang, and W.-Y. Zhou, ``Jittor: a novel
  deep learning framework with meta-operators and unified graph execution,''
  \emph{Science China Information Sciences}, vol.~63, no. 222103, pp. 1--21,
  2020.

\end{thebibliography}
}

\end{document}